\documentclass{egpubl}
\usepackage{eg2025}

\ConferenceSubmission

\usepackage[T1]{fontenc}
\usepackage{dfadobe}  

\usepackage{cite} 

\BibtexOrBiblatex
\electronicVersion
\PrintedOrElectronic

\ifpdf \usepackage[pdftex]{graphicx} \pdfcompresslevel=9
\else \usepackage[dvips]{graphicx} \fi

\usepackage{egweblnk} 
\usepackage{algorithm} 
\usepackage{algpseudocode} 
\usepackage{booktabs}
\usepackage{enumitem}
\usepackage{amsmath}

\usepackage{color}
\definecolor{blue}{rgb}{0,0,1}
\definecolor{red}{rgb}{1,0,0}
\definecolor{green}{rgb}{0,.5,0}
\definecolor{orange}{rgb}{0.75, 0.4, 0}
\definecolor{teal}{rgb}{0.0, 0.4, 0.4}
\definecolor{purple}{rgb}{0.65,0,0.65}
\definecolor{lightblue}{rgb}{0.52, 0.75,0.91}

\title{By-Example Synthesis of Vector Textures}
\author{}
\date{August 2024}

\author[C. Palazzolo \& O. van Kaick \& D. Mould]
{\parbox{\textwidth}{\centering C. Palazzolo$^{1}$, O. van Kaick$^{1}$ \& D. Mould$^{1}$
        }
        \\
{\parbox{\textwidth}{\centering $^1$Carleton University, Canada\\
       }
}
}

\begin{document}

\teaser{
 \includegraphics[width=1\linewidth]{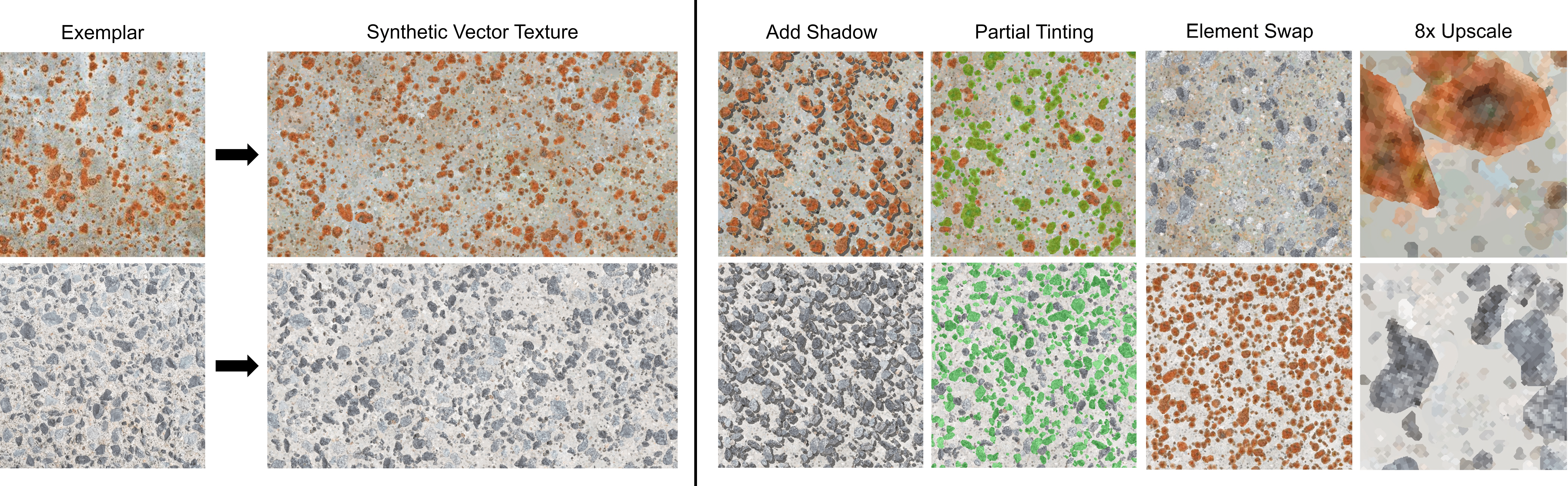}
 \centering
  \caption{
 Left: vector texture synthesis from raster input. Right: post-processing operations facilitated by our vector representation.
}
\label{fig:teaser}
}

\maketitle

\begin{abstract}
 We propose a new method for synthesizing an arbitrarily sized novel vector texture given a single raster exemplar. Our method first segments the exemplar to extract the primary textons, and then clusters them based on visual similarity. We then compute a descriptor to capture each texton's neighborhood which contains the inter-category relationships that are used at synthesis time. Next, we use a simple procedure to both extract and place the secondary textons behind the primary polygons. Finally, our method constructs a gradient field for the background which is defined by a set of data points and colors. The color of the secondary polygons are also adjusted to better match the gradient field. To compare our work with other methods, we use a wide range of perceptual-based metrics.



\end{abstract}  

\section{Introduction}



Texture synthesis is a long-studied problem, with raster images receiving a great deal of attention. Vector textures, however, are much less explored.
One advantage of vector textures is editability~\cite{diffusionCurves}: numerous existing operations can be used to manipulate~\cite{CSGOps} and remove elements from a vector texture. Such operations are much more difficult to apply to raster images. Figure~\ref{fig:teaser} shows two examples of synthesized vector textures along with a number of editing operations that are easily applied to polygons, but much harder to apply to a raster image.

Previous work has introduced methods for synthesizing simple polygon distributions through polygon packing~\cite{RepulsionPak,LearningGradientFields} or synthesis of vector patterns from exemplars~\cite{clusterVectorTexture,ContinuousCurveTextures,AppearanceGuidedSynthesis,PatchBasedGeometricTextureSynthesis}. However, these methods either take simple polygons as input or require the input to be already in vector format with separated vector elements. Other work explores generating vector content that matches a given raster image, but does not permit extrapolating a given exemplar image into a larger texture of a more stochastic nature~\cite{DVG}.

One might imagine generating vector textures from raster input by first transforming the raster input into vector format with existing tools~\cite{sun07,DVG}, then applying the methods discussed above to generate a new vector texture from the input. However, such an approach is likely to fail: vectorization tools often create output that is not separable, with many overlapping polygons exploiting transparency to recreate the input.


We propose a novel algorithm for synthesizing vector textures from a given exemplar in raster format (Figure~\ref{fig:teaser}). Our method is aimed at textures composed of distinct, irregular textons with uneven spacing and lacking global structure, e.g., rust and heterogeneous materials (Figure~\ref{fig:teaser}). In addition, we assume that the textures lack depth, implying absence of shadows and occlusions. From this input, we generate textures
composed of solid-colored polygons on a background gradient field. 

To synthesize a texture from an input raster exemplar, we build a texton hierarchy through segmentation and extract a background gradient field on which the textons reside. The texton hierarchy can be used to synthesize a novel vector texture of any size. We synthesize a background gradient over a target output domain, and then place textons over the background to replicate a similar texton context as in the exemplar. In addition, we slightly modify polygons to add more diversity to the output. 

We demonstrate with a set of experiments that our method is able to create textures in vector format that highly resemble the input exemplars, and is competitive when compared to methods that operate directly on the raster domain. Finally, we give examples of how the vector output facilitates editing operations on the textures that are difficult to accomplish with raster images, showing the advantage of the direct synthesis of vector textures from raster images.

In summary, this paper makes the following contributions:
\begin{itemize}
    \item We describe a method to convert a natural raster image to a vector representation containing discrete textons and a background gradient.
    \item We propose a method to synthesize novel vector textures given our vector representation.
    \item We demonstrate proof-of-concept editing operations made possible by our discrete texton representation. 
\end{itemize}

\section{Related Work}
To our knowledge, no existing methods directly synthesize a vector texture from a natural raster image. We consider natural textures, i.e., irregular textures with few or no repeated elements. Existing work on vector texture synthesis applies to line drawings and textures that are already in vector format, often with distributions of identical elements. Our method converts a natural image to vector format and then synthesizes a novel vector texture from it.

\textbf{Vector Texture Based Methods}. 
Both Tu et al.~\shortcite{clusterVectorTexture,ContinuousCurveTextures} and Hurtut et al.~\shortcite{AppearanceGuidedSynthesis} use polygons to produce novel textures, but these works assume the example texture is already comprised of polygons (for example, as an SVG file). The objective of these papers is clustering. Tu et al. group visually similar textons, whereas Hurtut et al. build clusters of samples for each vector element using local neighborhoods. Both methods attempt to identify the underlying local distribution of clusters and use this to synthesize novel textures which are visually similar. 
AlMeraj et al.~\cite{PatchBasedGeometricTextureSynthesis} propose an equivalent to the raster patch-based texture synthesis for vector images, also assuming input primitives are provided in vector form.

Saputra et al.~\cite{RepulsionPak} and Xue et al.~\cite{LearningGradientFields} describe vector texture synthesis as a polygon packing problem. These methods try to arrange a set of polygons in a finite area while avoiding overlap and reducing empty space.
These methods were designed for simple polygon distributions, and may not perform well on a vectorized natural image.

Methods also exist for vectorizing raster images. Sun et al.~\cite{DiffusionCurveTexture} describe a scale-independent explicit representation of a diffusion curve image that can be used for mapping textures onto a surface. Li et al.~\cite{DVG} propose an algorithm to  refine a set of vector graphics primitives to better match a raster image. The output of the method approximates well the input but is composed of overlapping transparent polygons with highly irregular shapes that lack semantic meaning. This makes it difficult to use the polygons for texture generation.


\textbf{Non-parametric Methods}.
Efros and Leung~\cite{PixelBasedSynthesis} propose a non-parametric sampling approach  where pixels are copied from the exemplar directly.
Later, Efros and Freeman~\cite{ImageQuilting} proposed Image Quilting, copying entire patches instead of individual pixels. Such approaches have extremely high quality outputs, but sometimes exhibit visible repetition. Our approach can be considered non-parametric owing to our texton reuse. We provide a quantitative comparison to Image Quilting.


Like our work, Galerne et al.~\cite{RandomPhaseTextures} and Heitz and Neyret~\cite{HistogramPreservingBlendingOperator} propose effective algorithms for synthesizing a narrow range of textures. Galerne et al.'s random-phase approach is designed for microtextures; Heitz and Neyret's Gaussianization method covers a similar class
of textures, although the texture types for which their method is successful
are not clearly characterized.

\textbf{Neural Network Based Methods}.
The literature on using neural networks for texture synthesis is very broad. Influential works such as Gatys et al.~\cite{Gatys2015}, Zhou et al.~\cite{TexExp}, and Jetchev et al.~\cite{SGAN} show how networks are able to create high resolution textures from a single smaller exemplar. Subsequent papers enhanced the ability of neural network-based synthesis through novel loss functions and architectures.

Bergmann et al.~\cite{PSGAN} propose the use of a Periodic Spatial GAN (PSGAN). This method improves on previous GAN-based texture synthesis approaches by using only convolutional layers in their network architectures. These improvements allow the method to learn textures that are periodic and of higher quality than results from previous GAN-based approaches. We compare with this method as representative of
the state of the are of GAN-based methods.

Zhou et al.~\cite{Zhou2023} propose a new loss function that combines Markov Random Fields with neural networks, the Guided Correspondence Loss (GCD Loss). Their approach is capable of synthesizing high-quality textures of arbitrary size. We use this method as a comparator due to its general effectiveness.

\textbf{Optimizer Based Methods}.
Kwatra et al.~\cite{TextureOptimization} propose a method for synthesizing novel, multi-scale textures through optimization. They introduce a global metric based on the Markov Random Field as the energy function and an optimization technique that is based on Expectation Maximization~\cite{EMOptimizer}. 
Kaspar et al.~\cite{SelfTuningOptimizer} 
present an example-based method capable of synthesizing high-quality textures, even those with nonstationary elements due to large-scale structures. We consider this method to be the
most effective optimization-based approach and include it as a basis for comparison.


\section{Vector Texture Synthesis}
 Our method takes a single stationary texture as input and produces vector output. 
The approach has two stages: an offline analysis, followed by an online synthesis step. 

In the analysis, we decompose the input texture into three layers:
a set of \textit{primary textons}, large and high-contrast elements
distinct from the background; \textit{secondary textons}, which are
smaller and less visually distinct than the primary textons;
and the remaining background region.
Then, we create a \textit{descriptor} for each primary texton, consisting of a local map of the other primary textons nearby. We estimate inter-element spacing
for the secondary textons, and we sample the background to obtain an
approximation of its color distribution.

Once the analysis phase is complete, the online synthesis step begins. Our method uses the descriptors and a scoring system for synthesizing the primary texton distribution, a simple point placement procedure that uses Poisson Disks~\cite{PoissonDisks} for the secondary elements, and data points for the background gradient field.

\begin{figure}
    \centering
    \includegraphics[width=1\linewidth]{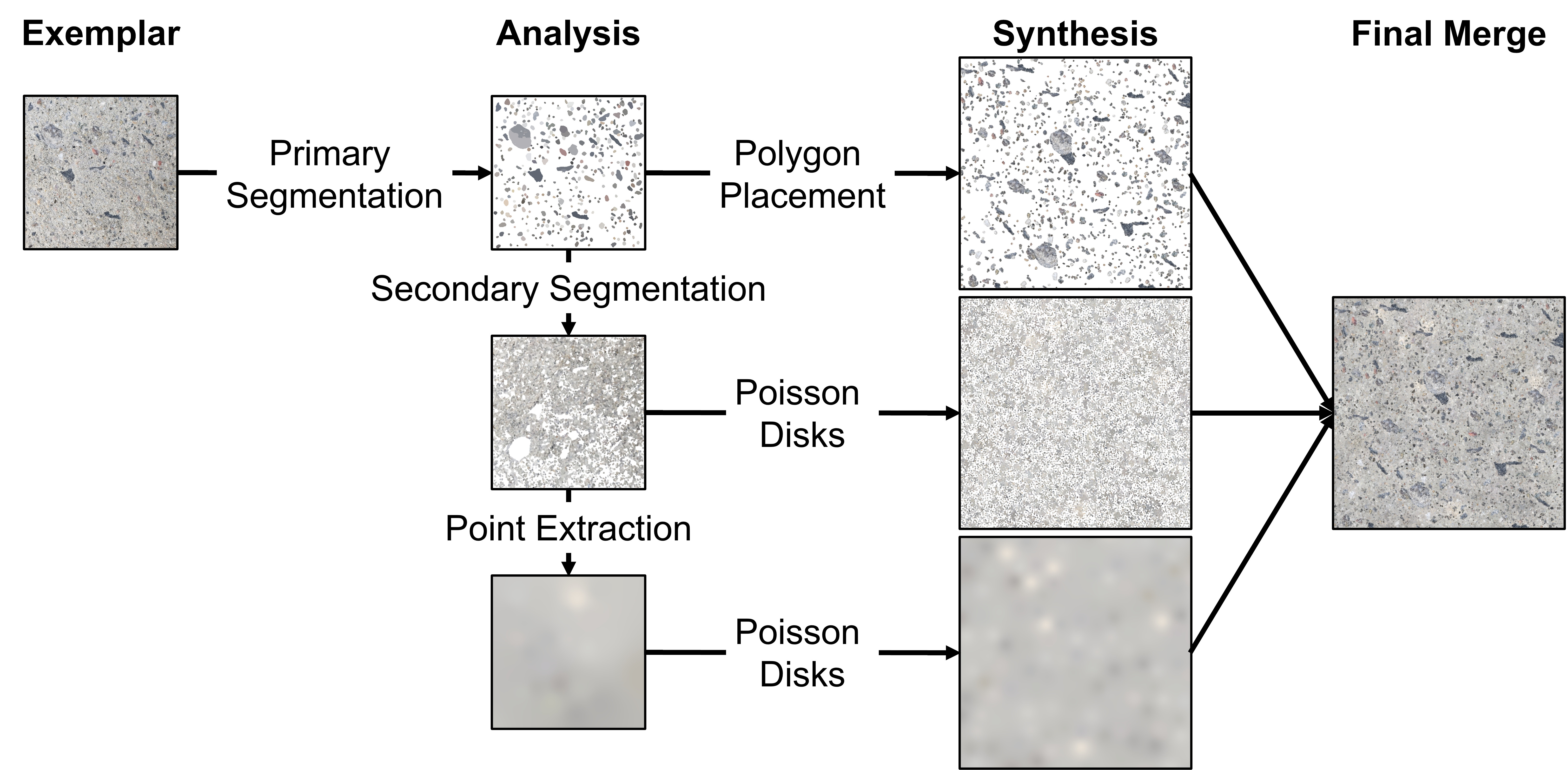}
    \caption{The high-level pipeline of our texture synthesis algorithm. 
    }
    \label{fig:pipeline}
\end{figure}

The overall pipeline is illustrated in Figure~\ref{fig:pipeline}.
Both the analysis and synthesis are described in greater detail in the following subsections.

\subsection{Analysis Phase}
\label{PolygonExtract}

The analysis phase has three main
subcomponents: first, identifying
major textons in the input; 
second, creating descriptors that
characterize the local spatial
arrangements of textons;
and third,
estimating secondary texton density
and background color gradient.
We discuss these three components in turn.

\subsubsection{Texton Extraction}
\label{foregroundPolygonExtraction}
We identify primary textons by segmenting the exemplar using Segment
Anything~\cite{SAM} with a high density of query points (10k on a 500$\times$500 image). The resulting set of segments are
processed to isolate individual connected
components and to remove duplicate textons.
This process
provides an initial set of textons
plus implicitly segments the
entire image into foreground (large textons) and background (area between
textons). Primary textons are subsegmented using a
flood-fill approach to obtain sub-texton detail.

\subsubsection{Secondary Texton Extraction}
\label{BackgroundInterElementSpacing}
Once the primary textons have been identified, we proceed to secondary textons. We segment secondary textons using floodfill from the remaining image; from these segments, we promote the largest (any whose area exceeds the median primary texton area from the previous step) to primary textons, typically around 5\% of them.

We then estimate the typical spacing between secondary textons by computing a Delaunay triangulation of the texton centres, then taking a predetermined percentile of the resulting edge lengths. We opted for the 40th percentile; empirically, the edge lengths are fairly stable over roughly the 30th to 60th percentiles, and a slightly lower estimate yields greater texton density in the synthesis and hence greater perceived detail in the output. This spacing estimate will be used in the synthesis stage.

\subsubsection{Primary Texton Descriptors}

The primary textons are clustered into categories. We first take the RGB color channels of the median color from any pixel contained within the texton, in addition to the texton area and Polsby-Popper compactness. These five numbers are used as the features of a K-Means Clustering model to create some number of clusters (we typically use 15). The three color features are weighted twice as heavily as the others.

\begin{figure}
    \includegraphics[width=.49\linewidth]{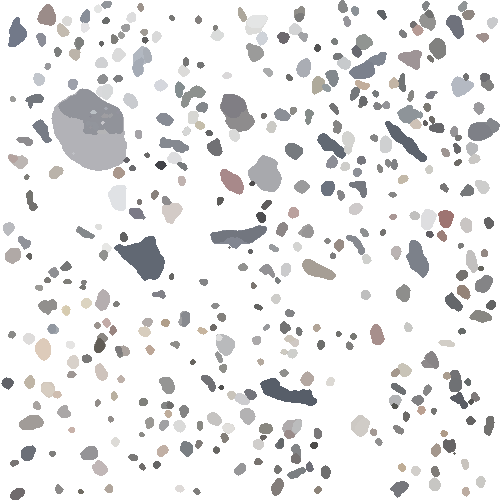}\hfill
    \includegraphics[width=.49\linewidth]{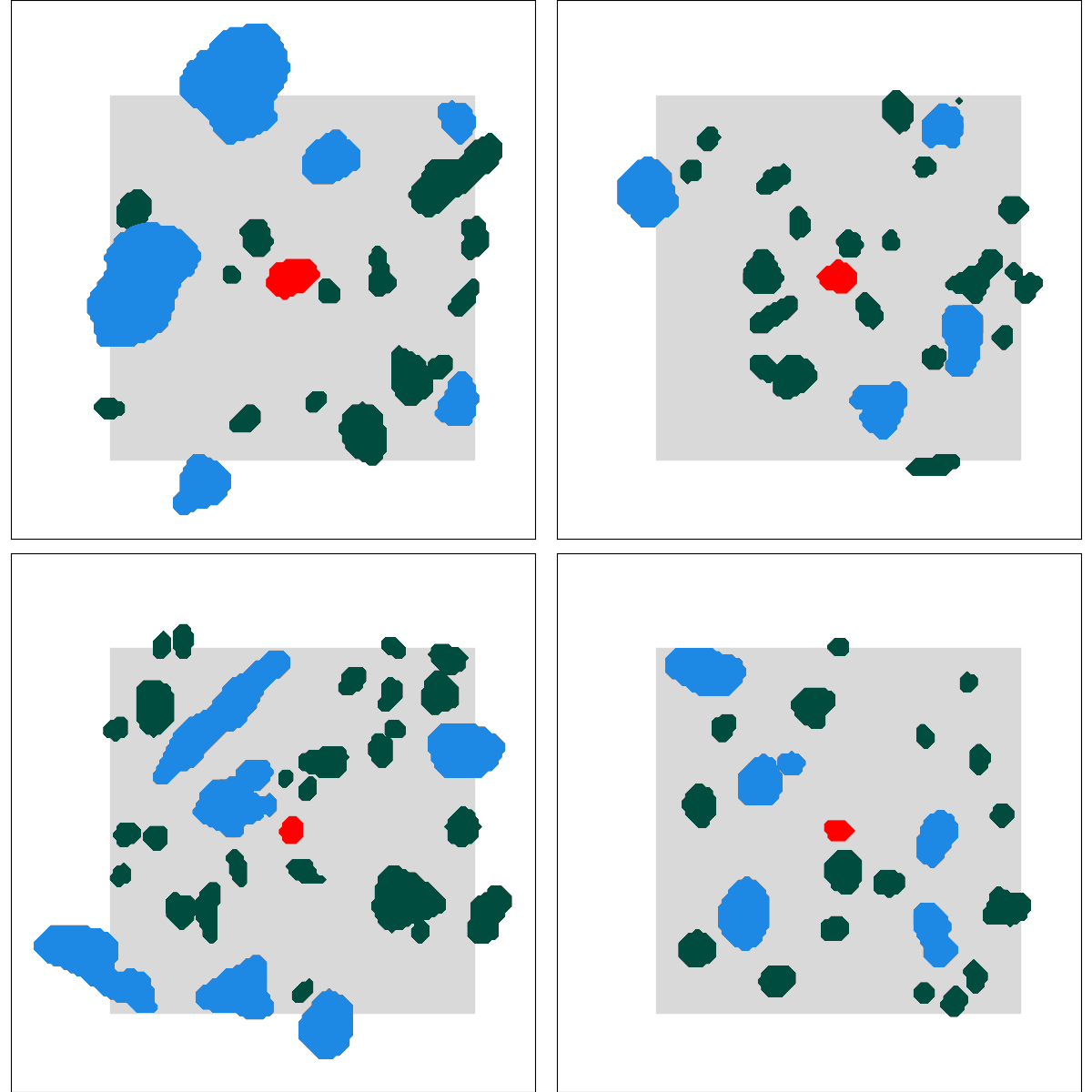}
    \caption{
    Illustration of descriptors extracted from a texture. Left: primary textons identified in the exemplar. Right: Four sample descriptors. The gray box indicates the boundary of the descriptor, the red polygon is the central texton, and the remaining polygons are the textons included in the descriptor, colored according to category. Notice that several polygons protrude from the initial descriptor boundary.
    }
    \label{fig:1ab}
\end{figure}

Our texton descriptor is a 2d grid-based map of the surroundings of a given texton, which we call the \textit{central texton}. Each cell of the map contains either a label for the category of the texton found there, if any, or a code for ``empty''. In addition, where there are textons that lie partially within the map and protrude beyond the initial boundaries, the map is extended to include these textons in full. Figure~\ref{fig:1ab} shows visualizations of some sample descriptors.

By default, we compute descriptors at the pixel resolution of the original raster input. However, we believe that smaller resolution would also work well, especially for future work in which more elaborate texton synthesis would be used, not necessarily matching the high-resolution texton shapes closely.

We require descriptors to be fully within the bounds of the image; descriptors extending outside the image are discarded. Thus, descriptors must be centred within the central region of the image, although they can include elements up to the edge of the image. In order to have enough variety of content,
our method requires large exemplars. 

\subsubsection{Background Gradient Field Summarization}
\label{BackgroundGradientFieldSummarization}
We next proceed to
characterizing the background. We remove from the starting image all pixels marked as part of any texton, then further remove an additional strip of pixels around each texton, ensuring that all remaining pixels strictly belong to the background.
We then construct a Voronoi diagram of a Poisson distribution of sites;
for each Voronoi region, we compute the median color of background pixels.
The set of median colors will be used to synthesize an output gradient field.

\subsection{Synthesis Phase}
To create a novel texture, we synthesize a distribution of primary and secondary textons along with a background gradient field. We then merge the resulting three layers into a single vector image. Details of synthesizing each layer appear in the following
subsections.

\subsubsection{Primary Texton Synthesis}
\label{foregroundSynth}

We aim to synthesize a spatial distribution of primary textons similar
to that in the exemplar. Akin to non-parametric sampling, we attempt to
match the surroundings of output textons with arrangements seen in the exemplar.

We track the desired output distribution through an incrementally-developed
target map, represented as a grid $G$. Initially all cells are coded ``undefined''.
Whenever we place a texton, we replace nearby undefined grid cells with target
values copied from the descriptor; values can be ``empty'' or can encode a
request for the cell to be covered by a texton of a given category.

\begin{figure}
    \centering
    \includegraphics[width=0.75\linewidth]{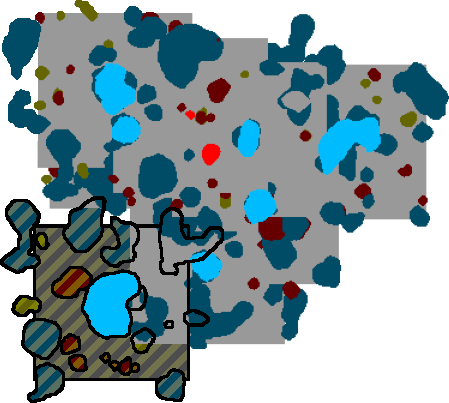}
    \caption{The map after different iterations of the synthesis process. White is unknown, light gray is empty, and blue, red, and yellow represent polygons. The newly added descriptor is highlighted with striping.}
    \label{fig:ProgressMap}
\end{figure}

The first texton is placed at random. Subsequent placements are made by trialing
additional textons and scoring each candidate, with the highest-scoring texton being
selected.
 After finding the best texton, we make an attempt to improve its location, repeatedly shifting it $+/-$ one grid cell in $x$ and $y$ and keeping the highest-scoring location.
 This texton's descriptor is then written onto the map; Figure~\ref{fig:ProgressMap} illustrates the process. 

Scoring is decided as follows. Call a candidate texton $C$; it will overlap with a region $R$ of the same category, say $\alpha$. 
The score for $C$ is a weighted average $\sum^{5}_{i=5}w_i A_i$;
we use weights $\textbf{w}=\left(0.5, -0.4, -0.2, -0.5, -0.5\right)$,
and the areas $A$ are computed by counting pixels $p$ with different properties, as follows:
\begin{enumerate}
 \item Target Area ($A_1$): $|| \{ p \in C \cap R \} || $ (overlap between texton and target region).
    
    \item Uncovered Area ($A_2$): $|| \{p \in R - C \} ||$ (area of target region left uncovered).
    \item Empty Area ($A_3$): $|| \{p \in C ~\text{such that}~ G[p]=\text{empty} \}||$ (texton covers an area that is supposed to be empty).
   
    \item Mismatched Area ($A_4$): $|| \{ p \in C ~\text{such that}~ G[p] != \{ \text{empty},\alpha \} \} || $ (texton overlaps an area where a different category is desired).
    
    \item Same Overlap Area ($A_5$)  $|| \{ p \in C - R ~\text{such that}~ G[p] = \alpha \} || $ (texton overlaps a different region of the same category).
\end{enumerate}


Algorithm~\ref{alg:foregroundSynth} describes the process for arranging the primary textons, and Algorithm~\ref{alg:polygonAccept} describes the process for adding a texton to the map.

\begin{algorithm}
	\caption{Primary Texton Layer} 
	\begin{algorithmic}[1]
        \State $G$ is the raster grid used for scoring (see text) 

        \item[]
        
        \State $\textbf{w}$, vector of weights for each area type
        \State $k$, number of tries to find the best placement per iteration
        \State $\varepsilon$, minimum placement score that can be accepted
        \State $F_{\max}$, maximum number of consecutive fails before the process is terminated

        \item[]
        
        \State $S$, the set of all textons and their descriptors extracted from the exemplar
        \State $f = 0$, number of consecutive failures

        \item[]

        \State Select a random texton and corresponding descriptor from $S$ as $T$ and $d$ respectively, weighted based on polygon area
        \State Add $T$ using Algorithm~\ref{alg:polygonAccept}

        \item[]

        \State // Iterate until the maximum number of consecutive failures has been exceeded
        \While{$f \leq F_{\max}$}        
            \State $P(x)$ is normalized probability distribution over possible texton placements

            \item[]

            \State // Try some textons and keep the one with the best score
            \State $b_s \leftarrow -\infty$, best placement score so far
        
            \For {$i = 1,2,\ldots,k$}            
                \State $c_c \leftarrow $ Sample $P(x)$ for random centroid pixel
                \State $c_t \leftarrow $ Random texton from $D$ with the category $G[c_c]$

                \item[]

                \State // Weighted sum to compute the score of the candidate polygon's placement
                \State $c_s \leftarrow \sum_{i=1}^{5}A_i\textbf{w}_i$  

                \If{$c_s > b_s$}
                    \State $b_s \leftarrow c_s$, $b_t \leftarrow c_t$, $b_c \leftarrow c_c$ 
                \EndIf
            \EndFor

            \item[]

%
%
%

            \State // If the best score is too low, reject the placement, otherwise add it
            \If{$b_s \leq \varepsilon$}
                \State $f \leftarrow f + 1$
                \State For all pixels $p \in b_t$, G[p] $\leftarrow$ empty
                \item[]
                
            \Else
                \State $f \leftarrow 0$
                \State Add $b_t$ at position $b_c$ using Algorithm~\ref{alg:polygonAccept}
            \EndIf

        \EndWhile
  
	\end{algorithmic} 
    \label{alg:foregroundSynth}
\end{algorithm}

\begin{algorithm}
	\caption{Texton Addition Procedure} 
	\begin{algorithmic}[1]
        \State $T$ = the added texton, $(a,b)$ = the location to add it, $d$ = its descriptor, $\alpha$ = its category

        \item[]

        \item // write placed texton into map
        \State For all pixels $p \in T$, $G[p] \leftarrow \alpha$.

        \item[]

        \item // copy descriptor to map wherever map was not yet defined
        \For{$(x, y)$ = all indices within $d$} 
            \If{$G[a+x, b+y]$ is undefined}
                \State $G[a+x, b+y] \leftarrow d[x, y]$
            \EndIf
        \EndFor   


	\end{algorithmic} 
    \label{alg:polygonAccept}
\end{algorithm}


\subsubsection{Secondary Texton Synthesis}
\label{BackgroundSynth}
Although background detail is needed for a convincing texture, the background is less complex and lower salience than the larger, higher-contrast primary textons, so it can be synthesized using a simpler method.
We use the mean inter-element spacing of secondary textons  (Section~\ref{BackgroundInterElementSpacing}) as the spacing for a Poisson disk distribution~\cite{PD85}. At each of these points, we place a random secondary texton. 

\subsubsection{Background Gradient Field}
\label{GradientMeshSynthesis}

In this step we synthesize a gradient field, interpolating between scattered points with associated color values. The process is to create a Poisson disk point distribution and assign to each point a randomly selected  color from the color pool extracted
in background analysis (Section~\ref{BackgroundGradientFieldSummarization}). The Poisson disk spacing is the same as that used for analysis.
 From these data points, the full field is computed using inverse-distance weighted interpolation~\cite{IDWInterpolation} with weights as $1/r^3$.

\subsection{Enhancements}
This section describes small changes to the base algorithm to improve the quality of results. All figures displayed in Section~\ref{results} use all of these improvements.

\textbf{Reduce duplication.} To ensure our method does not keep selecting the same textons repeatedly, whenever a texton is added to the output, we halve the probability of selecting that texton in future steps.


\textbf{Background matching.} When computing the median color of a secondary texton, we also compute a color delta against the background. 
At synthesis time, we determine the texton's color by adding the delta to the background color at the texton centroid. This small change improves results considerably.

\textbf{Global density correction.}
Our method tends to place more textons than necessary, so we remove some:
for each texton category, we repeatedly delete a randomly selected texton from that category until the fractional area covered by the category drops below the
exemplar's fractional area. 

\textbf{Smoother Gradient Field Interpolation.} We initially used $\frac{1}{r^3}$ for our scattered data interpolation, where $r$ is the distance between the coordinate and data point. However, we can get a smoother result by eliminating singularities. We instead suggest $\frac{1}{(r + r_0/4)^3}$, where $r_0$ is the spacing used to generate the Poisson disk distributions in both analysis and synthesis.

\section{Results and Discussion}
\label{results}
This section gives sample results and compares our synthetic textures to results generated by previous approaches.
We also give examples of
texture editing operations enabled by our vector representation, and discuss some limitations and failure cases.
Additional results and comparisons appear in the supplemental material.

Our approach is aimed at irregular natural images containing distinct textons, for which a vector representation of individual textons is beneficial. Ideally, the exemplars should be statistically stationary, 
the common assumption to virtually all example-based texture synthesis methods; we can somewhat cope with nonstationary backgrounds, though large-scale structures will not be preserved. Texton size, shape, and color can be heterogeneous and texton distribution, while stationary, should not follow any regular arrangement. Many natural textures possess these properties; a few examples appear in the figures of this section.

\begin{figure*}
    \centering
    \includegraphics[width=1\linewidth]{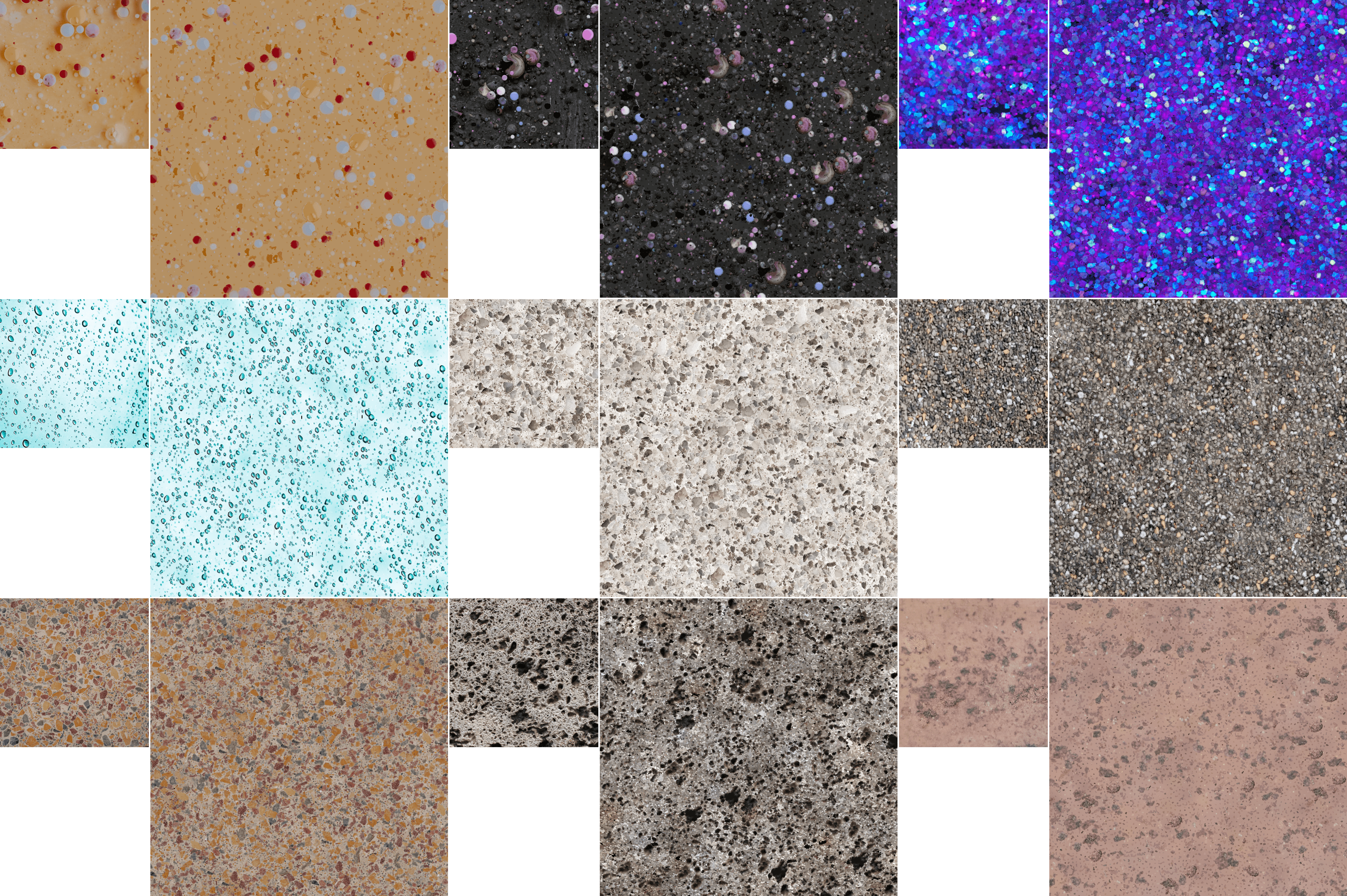}
    \caption{Textures synthesized using our algorithm. Each texture pair shows a raster exemplar (left, $500 \times 500$) and a synthetic vector image (right, rendered at $1000 \times 1000$).}
    \label{fig:ResultSample}
\end{figure*}

\subsection{Qualitative evaluation}
Figure~\ref{fig:ResultSample} shows textures synthesized by our method.  
Our method is capable of capturing and reproducing complex relationships between texture elements.

The top left image of Figure~\ref{fig:ResultSample} shows an exemplar with a few clusters of large, circular textons. Similar clusters appear in the result. This result appears flatter than the exemplar, since the primary texton segmentation did not capture shadows; shadows were instead identified as secondary textons and appear scattered throughout the background.


The centre row of images contains a mix of more distinct (left, right) and less distinct (centre) textons. In all cases, the textons are heterogeneous in size and shape. The vector output well approximates the style of the input without noticeable repetition.

The images in the bottom row contain distinct textons with irregular structure and spacing. These are the sort of exemplars we designed the method for, and in our view, these results are among the most convincingly natural of the images shown.

\subsection{Quantitative evaluation}
\begin{figure*}
    \centering
    \includegraphics[width=0.75\linewidth]{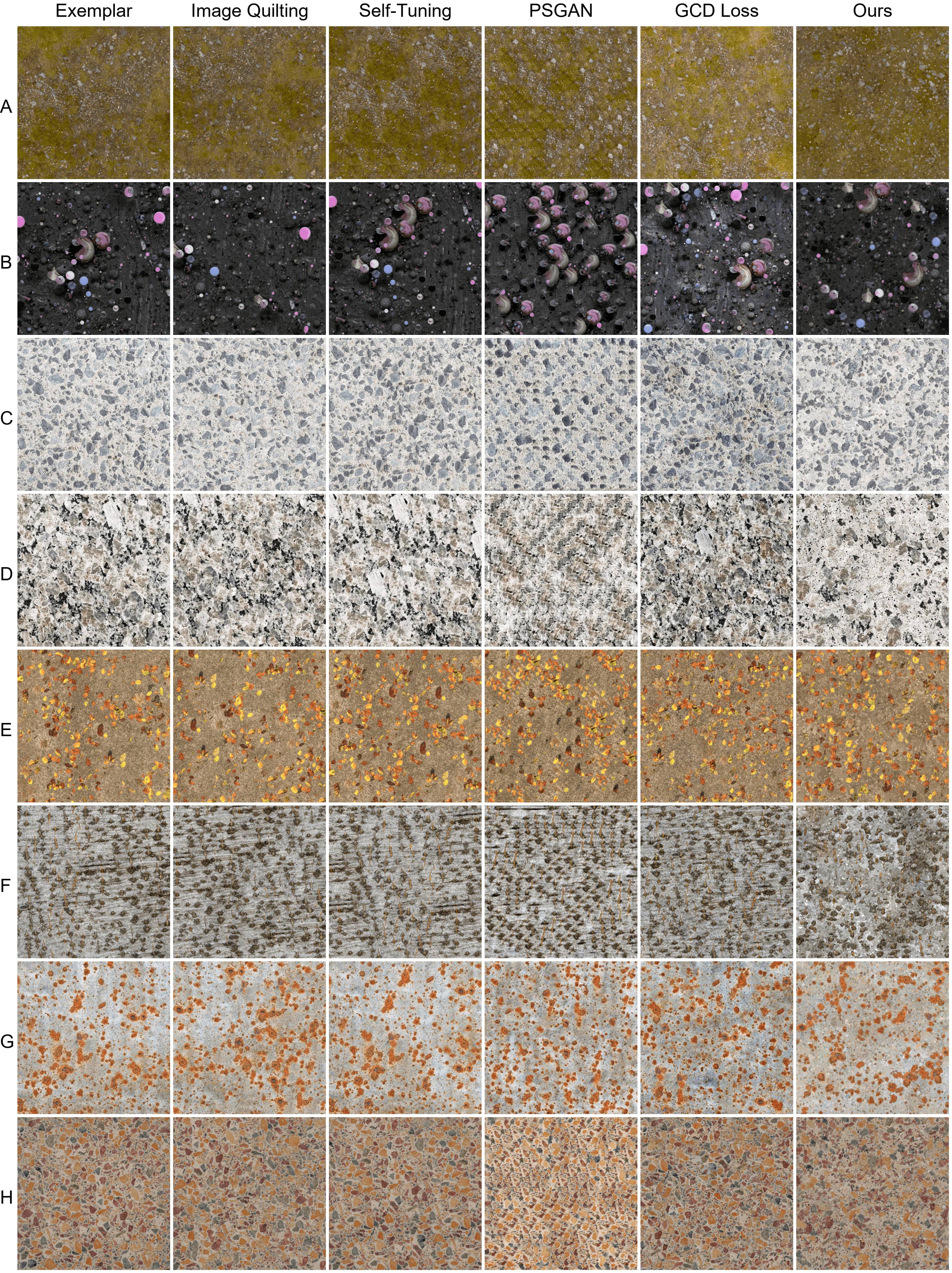}
    \caption{A comparison of our method to Image Quilting~\cite{ImageQuilting}, Self-Tuning Optimization~\cite{SelfTuningOptimizer}, PSGAN~\cite{PSGAN}, and GCD Loss~\cite{Zhou2023}. Default parameters were used. 
    }
    \label{fig:comparisonImages}
\end{figure*}

We report metrics comparing our results with those of selected previous synthesis methods that output textures in raster format: Image Quilting~\cite{ImageQuilting}, PSGAN~\cite{PSGAN}, GCD Loss~\cite{Zhou2023}, and Self-Tuning Optimization~\cite{SelfTuningOptimizer}. We use the authors' implementation whenever possible and use the default parameters. For Image Quilting, we use a patch size that is half the descriptor size used by our method. 
Sample outputs from these methods and ours are shown in Figure~\ref{fig:comparisonImages}.

We further compare our method to that of Tu et al.~\cite{clusterVectorTexture}, which processes textures in vector format. While both methods categorize elements based on visual similarity, the biggest difference is that Tu et al. has low diversity in each cluster, whereas we are not afforded that assumption due to working with natural images. As a result, our methods are not directly comparable, but we still include the comparison for completeness. For this comparison, we use five clusters of textons instead of 15, and set the uncovered area weight to 0.

Like Rodríguez-Pardo et al.~\cite{TexTile}, our evaluation uses the metrics SIFID~\cite{SinGAN}, CLIP-IQA~\cite{CLIP_IQA}, DISTS~\cite{ImageQualityAssessment}, PieAPP~\cite{PieAPP}, LPIPS~\cite{LPIPS}, and BRISQUE~\cite{BRISQUE} where each metric is reported to three decimal places. We use the implementation of these metrics provided by the PyIQA Python library~\cite{pyiqa}. 
We also compare the pixel intensity histograms between the exemplar and synthetic texture, reporting the Earth Mover's Distance  between the two.
Overall, these metrics fall into two categories: metrics that measure similarity and perceptual similarity (LPIPS, PieAPP, SIFID, EMD), and
metrics that measure image quality (BRISQUE, DISTS, Clip-IQA). Both fidelity to the exemplar and output image quality are important in texture synthesis applications.



We report a variety of quantitative metrics in Table~\ref{tab:comparisonTable}. Our method is competitive with state-of-the-art methods, with comparable scores in general and among the top  SIFID scores. We emphasize that our goal here is not to create another conventional raster-based texture synthesis method, but to work with vector textures and to be able to represent and manipulate individual textons.

\begin{figure}
    \centering
    \includegraphics[width=1\linewidth]{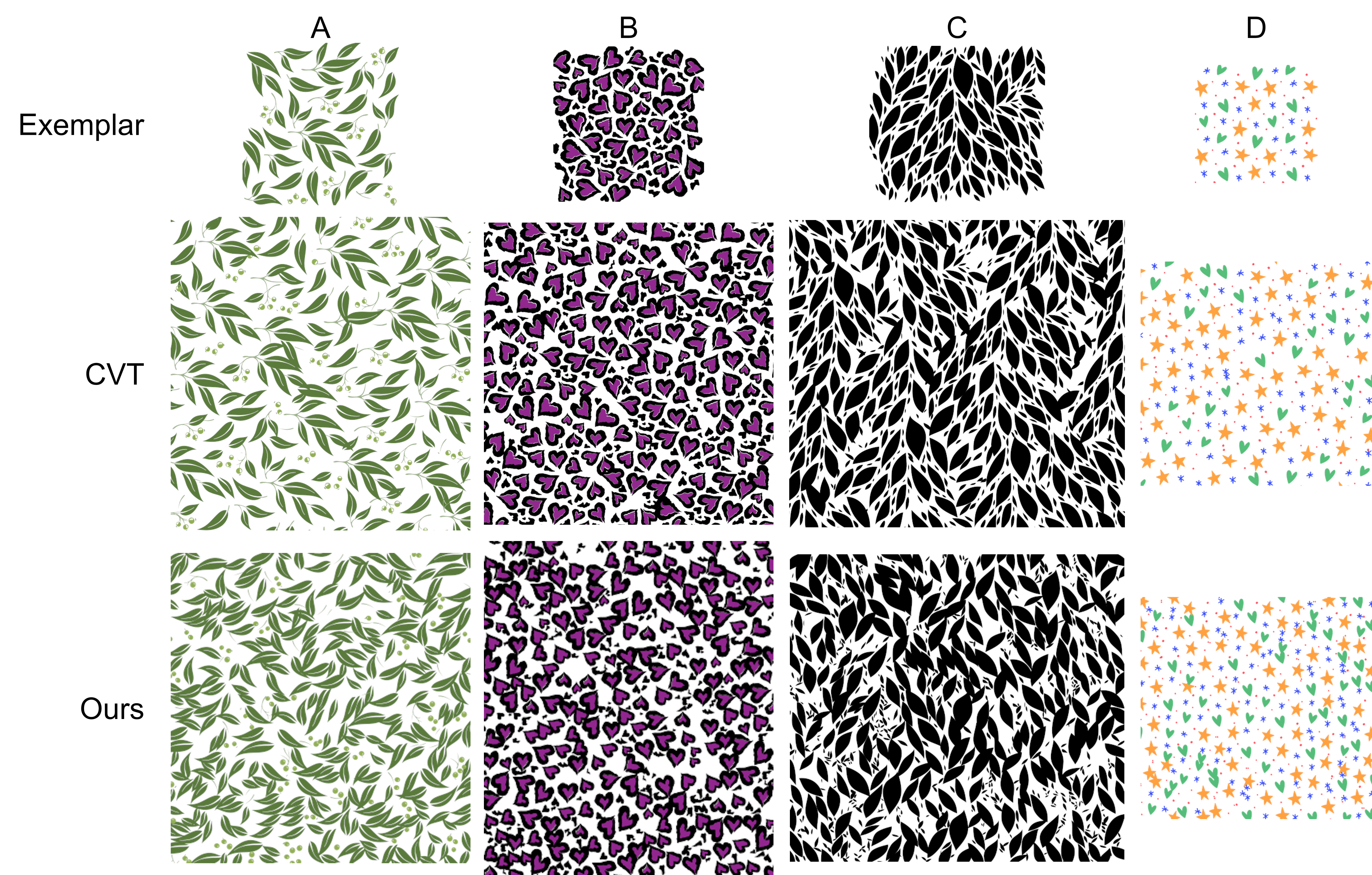}
    \caption{A comparison of our method to Tu et al.~\cite{clusterVectorTexture} on four vector texture exemplars. 
    }
    \label{fig:cvtComparison}
\end{figure}

To compare with the work of Tu et al.~\cite{clusterVectorTexture} which uses vector graphics, we convert all SVG images to raster with Inkscape and then use the perceptual similarity metrics, replacing transparent pixels with white color so that the metrics correctly consider the background of the textures. We also measure the fraction of image area covered by textons to get a coverage score. 
 Quantitative results are given in Table~\ref{tab:cvtComparisonMetrics},
 with  sample outputs in Figure~\ref{fig:cvtComparison}.
 From the scores, and from visual inspection of results, it is clear that 
 Tu et al.'s method is better able than our method to capture the target density of textons. Our method scores highly according to PieAPP and SIFID. Figure~\ref{fig:cvtComparison}b and c are composed of tightly packed vector elements; our method was not designed for densely packed and overlapping  textons and it struggles here. Further, these exemplars contain semi-structured arrangements that our method does not attempt to replicate. Despite this, our method still produces somewhat plausible results.

\newcommand{\TableLine}{\rule{0.75cm}{0.4pt}}

\begin{table*}
    \centering
\begin{tabular}{lrrrrr}
\hline
 Metric                    &        Image Quilting &        Self-Tuning &           PSGAN &             GCD Loss &          Ours \\
\hline
  LPIPS ↓                   &           \textit{0.486} &  \textbf{0.467} &           0.494 &           0.522 &           0.527 \\
 PieAPP ↓                  &  \textbf{1.063} &           1.129 &           1.239 &           1.087 &           \textit{1.072} \\
 SIFID ↓                   &  \textbf{0.000} &  \textbf{0.000} &  \textbf{0.000} &          \textit{18.634} &  \textbf{0.000} \\
 EMD ↓ &  \textbf{0.000} &  \textbf{0.000} &           \textit{0.001} &           \textit{0.001} &           \textit{0.001} \\
 \hline
 BRISQUE ↓       &          \textit{20.665} &          19.400 & \textbf{15.217} &          88.893 &          70.416 \\
 DISTS ↓                   &           \textit{0.144} &  \textbf{0.131} &           0.210 &           0.246 &           0.318 \\
 CLIP-IQA ↑      &           0.447 &  \textbf{0.479} &           \textit{0.459} &           0.405 &           0.441 \\
\hline
\end{tabular}
    \caption{Metrics computed for the images in Figure~\ref{fig:comparisonImages}, rounded to three decimal places; best scores are highlighted in bold and second best are italicized. Values are the mean of four results from each of the eight exemplars; per-image numbers appear in the supplementary material. The average exemplar BRISQUE score is 19.960 and the average exemplar CLIP-IQA score is 0.462. }
    \label{tab:comparisonTable}
\end{table*}

\begin{table}
    \centering
\begin{tabular}{lrr}
\hline
 Metric                    &      Tu et al. &           Ours \\
\hline
 LPIPS ↓                   & \textbf{0.498} &          0.508 \\
 PieAPP ↓                  &          2.856 & \textbf{2.803} \\
 SIFID ↓                   &          0.062 & \textbf{0.017} \\
 EMD ↓                     & \textbf{0.000} &          0.001 \\
 Coverage                  &          0.546 &          0.463 \\
 Absolute Difference ↓     & \textbf{0.011} &          0.083 \\
\hline
\end{tabular}    \caption{Metrics comparing our results with those of Tu et al.~\cite{clusterVectorTexture}.  The average coverage of the exemplars is 0.546; ``Absolute Difference'' is the mean of the pairwise differences between the synthetic image's coverage and the exemplar's coverage.}
    \label{tab:cvtComparisonMetrics}
\end{table}

\subsection{Ablation Study}
We conducted an ablation study to confirm the usefulness of the different components of our method.
We synthesized textures with various portions of the algorithm disabled: (A) solid background instead of a gradient field; (B) no updates to secondary texton color based on the gradient field; (C) texton placement probability is uniform, rather than informed by distance to the region perimeter; (D) no refinement steps; (E) no texton reselection penalty. When added to the full algorithm, modifications A-E provide six conditions to compare.
Table~\ref{tab:AblationStudy} contains a summary of the metrics computed over eight textures for each condition. All individual images as well as the full table of metrics can be found in the supplemental material. 

Based on the metrics, E is the preferred method. However, visual inspection of the figures (see supplemental material) reveals that E's textures contain substantial repetition, not considered by the metrics but which significantly reduces plausibility. 
The full algorithm is  best or second best after E in most of the similarity metrics.

\begin{table*}
    \centering
\begin{tabular}{lrrrrrr}
\hline
 Metric                    &         Full &      A &         B &   C &    D &         E \\
\hline
 LPIPS ↓                   &           0.519 &           0.520 &           0.520 &  \textbf{0.514} &           0.520 &           \textit{0.516} \\
 PieAPP ↓                  &           0.987 &           0.975 &           \textit{0.969} &           0.980 &           \textit{0.969} &  \textbf{0.930} \\
 SIFID ↓                   &  \textbf{0.042} &  \textbf{0.042} &  \textbf{0.042} &  \textbf{0.042} &  \textbf{0.042} &  \textbf{0.042} \\
 EMD ↓ &  \textbf{0.001} &           \textit{0.002} &  \textbf{0.001} &  \textbf{0.001} &  \textbf{0.001} &  \textbf{0.001} \\
 \hline
 BRISQUE ↓       &          67.318 &          68.463 & \textbf{66.877} &          67.205 &          68.559 &          \textit{66.974} \\
 DISTS ↓                   &           \textit{0.293} &           0.295 &           \textit{0.293} &           0.297 &           0.301 &  \textbf{0.291} \\
 CLIP-IQA ↑      &           \textit{0.548} &           \textit{0.548} &           0.551 &           0.552 &  \textbf{0.556} &           \textit{0.548} \\
\hline
\end{tabular}\caption{Ablation study metrics. Best scores are highlighted with bold, second best with italics. A: No background gradient field. B: Unmodified secondary texton color. C: Uniform starting point placement. D: No refinement steps. E: No texton repetition penalty. The average exemplar BRISQUE score is 19.960, and the average exemplar CLIP-IQA score is 0.462.}
    \label{tab:AblationStudy}
\end{table*}

\subsection{Timing and statistics}
Table~\ref{tab:BenchmarksTable} reports the average amount of time required to synthesize 4 $500 \times 500$ exemplars. 
Our method requires an average of 11.5 minutes for analysis.
We then take approximately 45 seconds to synthesize an exemplar, of which 3 seconds are required for the secondary textons and rasterizing the background gradient field. Synthesizing a larger $1000 \times 1000$ image takes about 4 minutes.

Table~\ref{tab:polygonCounts} shows texton counts after analyzing sample $500 \times 500$ exemplars.
A synthetic result of the same size contains an average of 479 primary and 4196 secondary textons, with 32656 detail polygons.


\begin{table}
    \centering
\begin{tabular}{lrr}
    \hline
    Method         & Analysis Time & Synthesis Time \\
    \hline
    Image Quilting$\dag$ & \TableLine & 21m \\
    Self-Tuning    & \TableLine & 38.5s\\
    PSGAN          & 7.5h & 0.5s \\
    GCD Loss       & \TableLine & 2.4m \\
    Ours           & 11.5m & 45s \\
    \hline
\end{tabular}
    \caption{The average time to synthesize a $500 \times 500$ exemplar. More than 60\% of our analysis time is due to SAM.
    $\dag${Image quilting results are from an unoptimized third-party implementation, not the author's code.}}
    \label{tab:BenchmarksTable}
\end{table}

\begin{table}
    \centering
    \begin{tabular}{lrrr}
        \hline
        Exemplar & Primary & Detail & Secondary \\
        \hline
        A & 1288 & 25039 & 3456 \\
        B & 874 & 33542 & 1308 \\
        C & 750 & 70505 & 5270 \\
        D & 312 & 54381 & 6602\\
        E & 596 & 77242 & 6674\\
        F & 490 & 69383 & 8802 \\
    \end{tabular}
    \caption{The number of primary and secondary textons in selected exemplars, plus the total number of detail polygons across all primary textons. Exemplar ID refers to the exemplars shown in Figure~\ref{fig:comparisonImages}. }
    \label{tab:polygonCounts}
\end{table}

\subsection{Vector Edits}
\begin{figure*}
    \centering
    \includegraphics[width=1\linewidth]{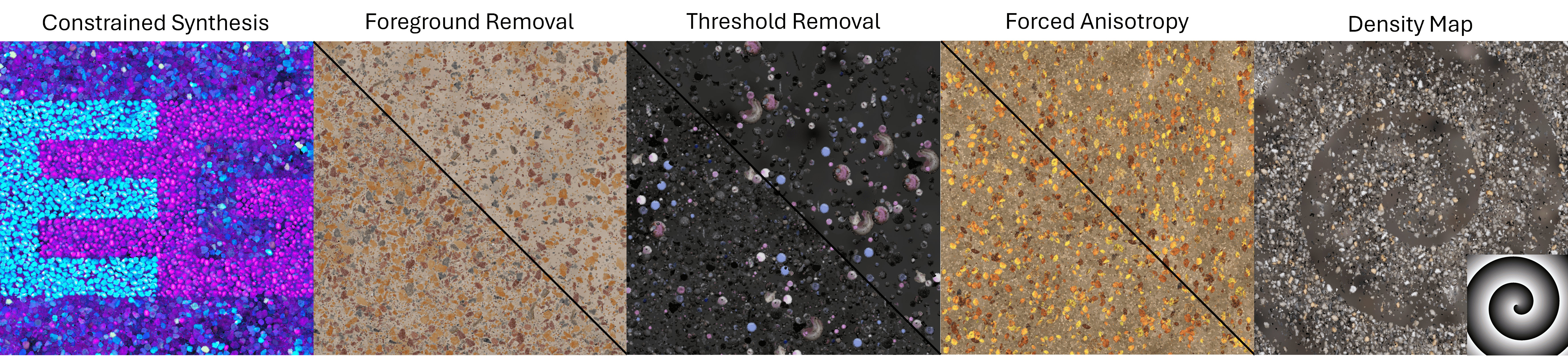}
    \caption{Some editing operations that are easy to perform given vector textons, but would be more difficult on a raster image. Additional editing operations are shown in Figure~\ref{fig:teaser}.}
    \label{fig:AdditionalEdits}
\end{figure*}

Results from a few post-processing operations were shown in Figure~\ref{fig:teaser}. Here, we give examples of some additional editing operations enabled by a vector representation. 
Element swap replaces each texton with a texton from a different image, maximizing overlap. Density map adjustment removes textons with probability proportional to the map intensity at that location. Forced anisotropy orients each texton along a desired global direction. Both random and threshold removal are texture simplification processes, removing textons to reduce the complexity and visual weight of the texture.
Figure~\ref{fig:AdditionalEdits} shows sample results from these processes.

\subsection{Limitations}
\label{limits}


Our method is intended for a limited class of textures: those with separate, identifiable textons, which most benefit from a vector representation. As individual textons become more difficult to distinguish, our method becomes less effective. Cases where textons overlap or are tightly packed also pose problems:  in such cases, texton shape and placement are dictated by nearby textons, whereas we assume more flexible arrangements to be possible.

We depend on image segmentation to identify vector elements, but segmentation is not always reliable. Textures without distinct elements will not yield segments and the subsequent synthesis will fail. Shadows may produce separate segments, and subsequently in synthesis can become detached from any possible source of shadow, with unappealing results.

\begin{figure}
    \centering
    \includegraphics[width=1\linewidth]{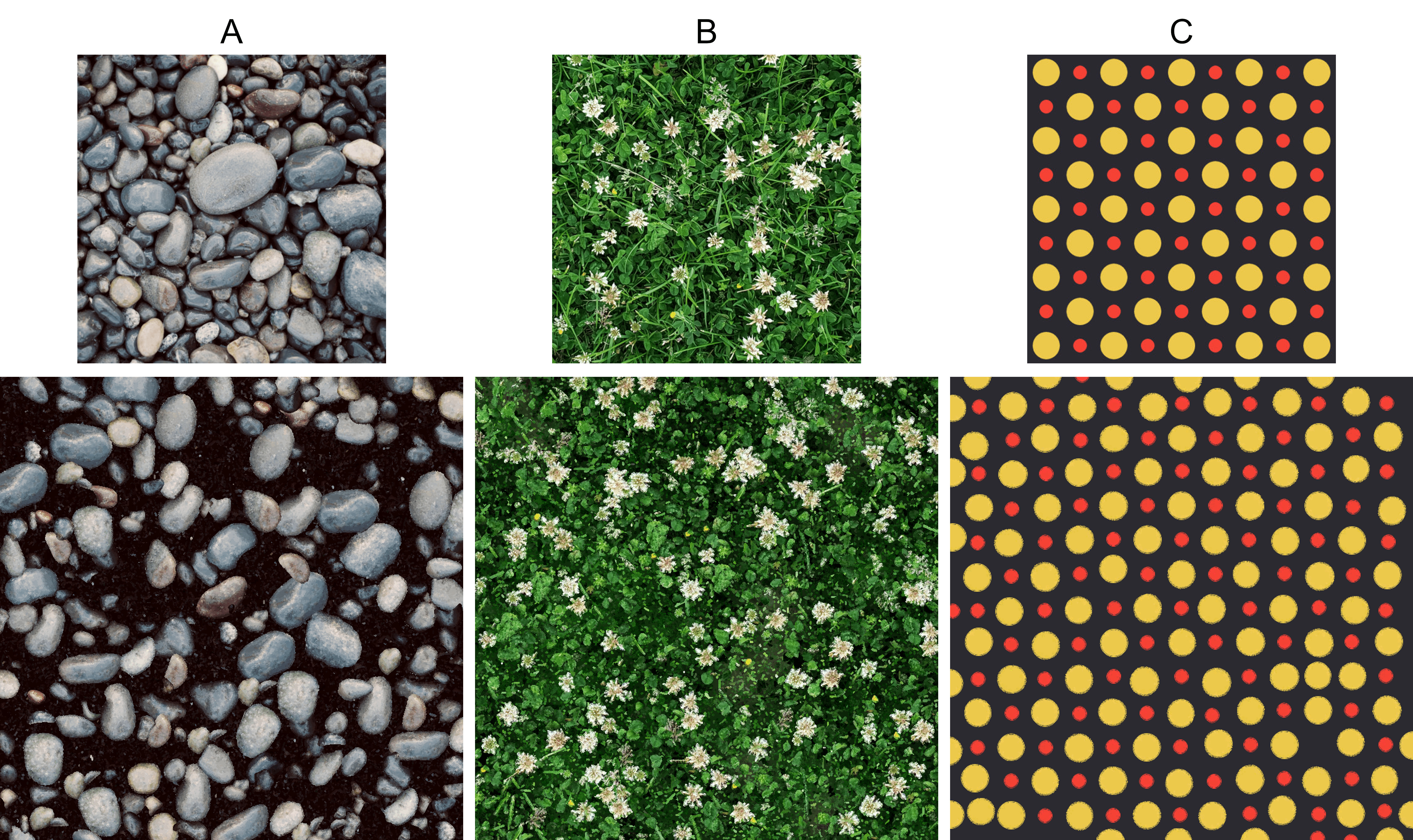}
    \caption{A few failure cases. Tight packing, semantic relationships and shadows, and highly regular structures are challenging. 
    }
    \label{fig:limitations}
\end{figure}

Figure~\ref{fig:limitations} shows the result of our method when used on some of these cases. A: A texture with occlusions yields irregular textons. B: Semantic relationships may not be preserved and objects may not be adequately separated through segmentation, degrading results. C: Uniform textures are generally handled adequately, but a few highly salient errors spoil the effect.

\section{Conclusion}

In this paper, we have proposed a novel algorithm for synthesizing vector textures composed of solid-colored polygons and a background gradient field. We start with an input raster exemplar, and convert it to a vector representation through image segmentation. We then take this representation to synthesize a novel vector texture. We also showcase some post-processing operations on our vector results that would be difficult to apply to a raster image, illustrating the advantage of vector textures.

We compare the results of our method with a number of other state-of-the-art algorithms in both a qualitative and quantitative analysis. Our method is competitive with the other algorithms despite our textures being purely composed of vector elements. We also compared our method with a vector texture synthesis algorithm designed for structured textures; despite the different types of textures the methods were designed for, our results are comparable.

There remain some opportunities for future work. Our polygons are solid-colored (sub-texton detail is provided by a hierarchy of polygons) and using gradients to enhance detail is a clear next step. Exploring alternatives to pure segmentation for polygon extraction could yield improvements; manual intervention or polygon repair could be investigated. Finally, reconstructing data other than color (e.g., normals) could be beneficial to some applications.

\bibliographystyle{eg-alpha-doi} 
\bibliography{egbibsample}   

\newcommand{\etalchar}[1]{$^{#1}$}
\begin{thebibliography}{\uppercase{RPCGLM24}}

\bibitem[AKA13]{PatchBasedGeometricTextureSynthesis}
\textsc{AlMeraj Z., Kaplan C.~S., Asente P.}:
\newblock Patch-based geometric texture synthesis.
\newblock In \emph{Proceedings of the Symposium on Computational Aesthetics} (New York, NY, USA, 2013), CAE '13, Association for Computing Machinery, p.~15–19.
\newblock URL: \url{https://doi.org/10.1145/2487276.2487278}, \href {https://doi.org/10.1145/2487276.2487278} {\path{doi:10.1145/2487276.2487278}}.

\bibitem[BJV17]{PSGAN}
\textsc{Bergmann U., Jetchev N., Vollgraf R.}:
\newblock Learning texture manifolds with the periodic spatial gan, 2017.
\newblock URL: \url{https://arxiv.org/abs/1705.06566}, \href {http://arxiv.org/abs/1705.06566} {\path{arXiv:1705.06566}}.

\bibitem[Bri07]{PoissonDisks}
\textsc{Bridson R.}:
\newblock Fast poisson disk sampling in arbitrary dimensions.
\newblock In \emph{ACM SIGGRAPH 2007 Sketches} (New York, NY, USA, 2007), SIGGRAPH '07, Association for Computing Machinery, p.~22–es.
\newblock URL: \url{https://doi.org/10.1145/1278780.1278807}, \href {https://doi.org/10.1145/1278780.1278807} {\path{doi:10.1145/1278780.1278807}}.

\bibitem[CM22]{pyiqa}
\textsc{Chen C., Mo J.}:
\newblock {IQA-PyTorch}: Pytorch toolbox for image quality assessment.
\newblock [Online]. Available: \url{https://github.com/chaofengc/IQA-PyTorch}, 2022.

\bibitem[DMWS20]{ImageQualityAssessment}
\textsc{Ding K., Ma K., Wang S., Simoncelli E.~P.}:
\newblock Image quality assessment: Unifying structure and texture similarity.
\newblock \emph{CoRR abs/2004.07728} (2020).
\newblock URL: \url{https://arxiv.org/abs/2004.07728}, \href {http://arxiv.org/abs/2004.07728} {\path{arXiv:2004.07728}}.

\bibitem[DW85]{PD85}
\textsc{Dipp\'{e} M. A.~Z., Wold E.~H.}:
\newblock Antialiasing through stochastic sampling.
\newblock \emph{SIGGRAPH Comput. Graph. 19}, 3 (July 1985), 69–78.
\newblock URL: \url{https://doi.org/10.1145/325165.325182}, \href {https://doi.org/10.1145/325165.325182} {\path{doi:10.1145/325165.325182}}.

\bibitem[EF01]{ImageQuilting}
\textsc{Efros A.~A., Freeman W.~T.}:
\newblock \emph{Image Quilting for Texture Synthesis and Transfer}, 1~ed.
\newblock Association for Computing Machinery, New York, NY, USA, 2001, pp.~571--576.
\newblock URL: \url{https://doi.org/10.1145/3596711.3596771}.

\bibitem[EL99]{PixelBasedSynthesis}
\textsc{Efros A., Leung T.}:
\newblock Texture synthesis by non-parametric sampling.
\newblock In \emph{Proceedings of the Seventh IEEE International Conference on Computer Vision} (1999), vol.~2, pp.~1033--1038 vol.2.
\newblock \href {https://doi.org/10.1109/ICCV.1999.790383} {\path{doi:10.1109/ICCV.1999.790383}}.

\bibitem[GEB15]{Gatys2015}
\textsc{Gatys L.~A., Ecker A.~S., Bethge M.}:
\newblock Texture synthesis using convolutional neural networks, 2015.
\newblock URL: \url{https://arxiv.org/abs/1505.07376}, \href {http://arxiv.org/abs/1505.07376} {\path{arXiv:1505.07376}}.

\bibitem[GGM11]{RandomPhaseTextures}
\textsc{Galerne B., Gousseau Y., Morel J.-M.}:
\newblock Random phase textures: Theory and synthesis.
\newblock \emph{IEEE Transactions on Image Processing 20}, 1 (2011), 257--267.
\newblock \href {https://doi.org/10.1109/TIP.2010.2052822} {\path{doi:10.1109/TIP.2010.2052822}}.

\bibitem[HLT{\etalchar{*}}09]{AppearanceGuidedSynthesis}
\textsc{Hurtut T., Landes P.-E., Thollot J., Gousseau Y., Drouillhet R., Coeurjolly J.-F.}:
\newblock Appearance-guided synthesis of element arrangements by example.
\newblock In \emph{Proceedings of the 7th International Symposium on Non-Photorealistic Animation and Rendering} (New York, NY, USA, 2009), NPAR '09, Association for Computing Machinery, p.~51–60.
\newblock URL: \url{https://doi.org/10.1145/1572614.1572623}, \href {https://doi.org/10.1145/1572614.1572623} {\path{doi:10.1145/1572614.1572623}}.

\bibitem[HN18]{HistogramPreservingBlendingOperator}
\textsc{Heitz E., Neyret F.}:
\newblock High-performance by-example noise using a histogram-preserving blending operator.
\newblock \emph{Proc. ACM Comput. Graph. Interact. Tech. 1}, 2 (aug 2018).
\newblock URL: \url{https://doi.org/10.1145/3233304}, \href {https://doi.org/10.1145/3233304} {\path{doi:10.1145/3233304}}.

\bibitem[JBV17]{SGAN}
\textsc{Jetchev N., Bergmann U., Vollgraf R.}:
\newblock Texture synthesis with spatial generative adversarial networks, 2017.
\newblock URL: \url{https://arxiv.org/abs/1611.08207}, \href {http://arxiv.org/abs/1611.08207} {\path{arXiv:1611.08207}}.

\bibitem[KEBK05]{TextureOptimization}
\textsc{Kwatra V., Essa I., Bobick A., Kwatra N.}:
\newblock Texture optimization for example-based synthesis.
\newblock In \emph{ACM SIGGRAPH 2005 Papers} (New York, NY, USA, 2005), SIGGRAPH '05, Association for Computing Machinery, p.~795–802.
\newblock URL: \url{https://doi.org/10.1145/1186822.1073263}, \href {https://doi.org/10.1145/1186822.1073263} {\path{doi:10.1145/1186822.1073263}}.

\bibitem[KMR{\etalchar{*}}23]{SAM}
\textsc{Kirillov A., Mintun E., Ravi N., Mao H., Rolland C., Gustafson L., Xiao T., Whitehead S., Berg A.~C., Lo W.-Y., Doll{\'a}r P., Girshick R.}:
\newblock Segment anything.
\newblock \emph{arXiv:2304.02643} (2023).

\bibitem[KNL{\etalchar{*}}15]{SelfTuningOptimizer}
\textsc{Kaspar A., Neubert B., Lischinski D., Pauly M., Kopf J.}:
\newblock Self tuning texture optimization.
\newblock \emph{Computer Graphics Forum 34}, 2 (2015), 349--359.
\newblock URL: \url{https://onlinelibrary.wiley.com/doi/abs/10.1111/cgf.12565}, \href {http://arxiv.org/abs/https://onlinelibrary.wiley.com/doi/pdf/10.1111/cgf.12565} {\path{arXiv:https://onlinelibrary.wiley.com/doi/pdf/10.1111/cgf.12565}}, \href {https://doi.org/https://doi.org/10.1111/cgf.12565} {\path{doi:https://doi.org/10.1111/cgf.12565}}.

\bibitem[LLGRK20]{DVG}
\textsc{Li T.-M., Luk\'{a}\v{c} M., Gharbi M., Ragan-Kelley J.}:
\newblock Differentiable vector graphics rasterization for editing and learning.
\newblock \emph{ACM Trans. Graph. 39}, 6 (Nov. 2020).
\newblock URL: \url{https://doi.org/10.1145/3414685.3417871}, \href {https://doi.org/10.1145/3414685.3417871} {\path{doi:10.1145/3414685.3417871}}.

\bibitem[LTH86]{CSGOps}
\textsc{Laidlaw D.~H., Trumbore W.~B., Hughes J.~F.}:
\newblock Constructive solid geometry for polyhedral objects.
\newblock \emph{SIGGRAPH Comput. Graph. 20}, 4 (aug 1986), 161–170.
\newblock URL: \url{https://doi.org/10.1145/15886.15904}, \href {https://doi.org/10.1145/15886.15904} {\path{doi:10.1145/15886.15904}}.

\bibitem[MK08]{EMOptimizer}
\textsc{McLachlan G.~J., Krishnan T.}:
\newblock \emph{The EM Algorithm and Extensions, 2E}.
\newblock Wiley Series in Probability and Statistics. John Wiley \& Sons, Inc., Hoboken, NJ, USA, Feb. 2008.
\newblock URL: \url{http://doi.wiley.com/10.1002/9780470191613}, \href {https://doi.org/10.1002/9780470191613} {\path{doi:10.1002/9780470191613}}.

\bibitem[MMB12]{BRISQUE}
\textsc{Mittal A., Moorthy A.~K., Bovik A.~C.}:
\newblock No-reference image quality assessment in the spatial domain.
\newblock \emph{IEEE Transactions on Image Processing 21}, 12 (2012), 4695--4708.
\newblock \href {https://doi.org/10.1109/TIP.2012.2214050} {\path{doi:10.1109/TIP.2012.2214050}}.

\bibitem[OBW{\etalchar{*}}08]{diffusionCurves}
\textsc{Orzan A., Bousseau A., Winnem\"{o}ller H., Barla P., Thollot J., Salesin D.}:
\newblock Diffusion curves: a vector representation for smooth-shaded images.
\newblock \emph{ACM Trans. Graph. 27}, 3 (aug 2008), 1–8.
\newblock URL: \url{https://doi.org/10.1145/1360612.1360691}, \href {https://doi.org/10.1145/1360612.1360691} {\path{doi:10.1145/1360612.1360691}}.

\bibitem[PCMS18]{PieAPP}
\textsc{Prashnani E., Cai H., Mostofi Y., Sen P.}:
\newblock Pieapp: Perceptual image-error assessment through pairwise preference, 2018.
\newblock URL: \url{https://arxiv.org/abs/1806.02067}, \href {http://arxiv.org/abs/1806.02067} {\path{arXiv:1806.02067}}.

\bibitem[RPCGLM24]{TexTile}
\textsc{Rodriguez-Pardo C., Casas D., Garces E., Lopez-Moreno J.}:
\newblock Textile: A differentiable metric for texture tileability.
\newblock In \emph{Proceedings of the IEEE/CVF Conference on Computer Vision and Pattern Recognition (CVPR)} (2024).

\bibitem[SDM19]{SinGAN}
\textsc{Shaham T.~R., Dekel T., Michaeli T.}:
\newblock Sin{G}{A}{N}: Learning a generative model from a single natural image.
\newblock In \emph{2019 IEEE/CVF International Conference on Computer Vision (ICCV)} (2019), pp.~4569--4579.
\newblock \href {https://doi.org/10.1109/ICCV.2019.00467} {\path{doi:10.1109/ICCV.2019.00467}}.

\bibitem[She68]{IDWInterpolation}
\textsc{Shepard D.}:
\newblock A two-dimensional interpolation function for irregularly-spaced data.
\newblock In \emph{Proceedings of the 1968 23rd ACM National Conference} (New York, NY, USA, 1968), ACM '68, Association for Computing Machinery, p.~517–524.
\newblock URL: \url{https://doi.org/10.1145/800186.810616}, \href {https://doi.org/10.1145/800186.810616} {\path{doi:10.1145/800186.810616}}.

\bibitem[SKA21]{RepulsionPak}
\textsc{Saputra R.~A., Kaplan C.~S., Asente P.}:
\newblock Improved deformation-driven element packing with repulsionpak.
\newblock \emph{IEEE Transactions on Visualization and Computer Graphics 27}, 4 (2021), 2396--2408.
\newblock \href {https://doi.org/10.1109/TVCG.2019.2950235} {\path{doi:10.1109/TVCG.2019.2950235}}.

\bibitem[SLWS07]{sun07}
\textsc{Sun J., Liang L., Wen F., Shum H.-Y.}:
\newblock Image vectorization using optimized gradient meshes.
\newblock \emph{ACM Transactions on Graphics (TOG) 26}, 3 (2007).

\bibitem[SXD{\etalchar{*}}12]{DiffusionCurveTexture}
\textsc{Sun X., Xie G., Dong Y., Lin S., Xu W., Wang W., Tong X., Guo B.}:
\newblock Diffusion curve textures for resolution independent texture mapping.
\newblock \emph{ACM Trans. Graph. 31}, 4 (July 2012).
\newblock URL: \url{https://doi-org.proxy.library.carleton.ca/10.1145/2185520.2185570}, \href {https://doi.org/10.1145/2185520.2185570} {\path{doi:10.1145/2185520.2185570}}.

\bibitem[TWY{\etalchar{*}}20]{ContinuousCurveTextures}
\textsc{Tu P., Wei L.-Y., Yatani K., Igarashi T., Zwicker M.}:
\newblock Continuous curve textures.
\newblock \emph{ACM Transactions on Graphics 39}, 6 (Nov. 2020), 1–16.
\newblock URL: \url{http://dx.doi.org/10.1145/3414685.3417780}, \href {https://doi.org/10.1145/3414685.3417780} {\path{doi:10.1145/3414685.3417780}}.

\bibitem[TWZ22]{clusterVectorTexture}
\textsc{Tu P., Wei L.-Y., Zwicker M.}:
\newblock Clustered vector textures.
\newblock \emph{ACM Trans. Graph. 41}, 4 (jul 2022).
\newblock URL: \url{https://doi.org/10.1145/3528223.3530062}, \href {https://doi.org/10.1145/3528223.3530062} {\path{doi:10.1145/3528223.3530062}}.

\bibitem[WCL23]{CLIP_IQA}
\textsc{Wang J., Chan K.~C., Loy C.~C.}:
\newblock Exploring {CLIP} for assessing the look and feel of images.
\newblock \emph{Proceedings of the AAAI Conference on Artificial Intelligence 37}, 2 (Jun. 2023), 2555--2563.
\newblock URL: \url{https://ojs.aaai.org/index.php/AAAI/article/view/25353}, \href {https://doi.org/10.1609/aaai.v37i2.25353} {\path{doi:10.1609/aaai.v37i2.25353}}.

\bibitem[XWL{\etalchar{*}}23]{LearningGradientFields}
\textsc{Xue T., Wu M., Lu L., Wang H., Dong H., Chen B.}:
\newblock Learning gradient fields for scalable and generalizable irregular packing, 2023.
\newblock URL: \url{https://arxiv.org/abs/2310.19814}, \href {http://arxiv.org/abs/2310.19814} {\path{arXiv:2310.19814}}.

\bibitem[ZCXH23]{Zhou2023}
\textsc{Zhou Y., Chen K., Xiao R., Huang H.}:
\newblock Neural texture synthesis with guided correspondence.
\newblock In \emph{Proceedings of the IEEE/CVF Conference on Computer Vision and Pattern Recognition (CVPR)} (June 2023), pp.~18095--18104.

\bibitem[ZIE{\etalchar{*}}18]{LPIPS}
\textsc{Zhang R., Isola P., Efros A.~A., Shechtman E., Wang O.}:
\newblock The unreasonable effectiveness of deep features as a perceptual metric.
\newblock In \emph{2018 IEEE/CVF Conference on Computer Vision and Pattern Recognition} (2018), pp.~586--595.
\newblock \href {https://doi.org/10.1109/CVPR.2018.00068} {\path{doi:10.1109/CVPR.2018.00068}}.

\bibitem[ZZB{\etalchar{*}}18]{TexExp}
\textsc{Zhou Y., Zhu Z., Bai X., Lischinski D., Cohen-Or D., Huang H.}:
\newblock Non-stationary texture synthesis by adversarial expansion.
\newblock \emph{ACM Transactions on Graphics 37}, 4 (July 2018), 1–13.
\newblock URL: \url{http://dx.doi.org/10.1145/3197517.3201285}, \href {https://doi.org/10.1145/3197517.3201285} {\path{doi:10.1145/3197517.3201285}}.

\end{thebibliography}

\end{document}